\pdfoutput=1

\documentclass[11pt]{article}

\usepackage{emnlp2021}
\usepackage{textcomp}

\usepackage{times}
\usepackage{latexsym}

\usepackage[T1]{fontenc}

\usepackage[utf8]{inputenc}

\usepackage{microtype}

\usepackage{listings}
\usepackage{xcolor}

\usepackage{multirow}
\usepackage{siunitx}

\usepackage{balance}

\definecolor{delim}{RGB}{20,105,176}
\definecolor{numb}{RGB}{106, 109, 32}
\definecolor{string}{rgb}{0.64,0.08,0.08}

\lstdefinelanguage{json}{
    numbers=left,
    numberstyle=\small,
    frame=single,
    rulecolor=\color{black},
    showspaces=false,
    showtabs=false,
    breaklines=true,
    breakatwhitespace=true,
    captionpos=b,
    basicstyle=\ttfamily\small,
    upquote=true,
    stringstyle=\color{string},
    literate=
     *{0}{{{\color{numb}0}}}{1}
      {1}{{{\color{numb}1}}}{1}
      {2}{{{\color{numb}2}}}{1}
      {3}{{{\color{numb}3}}}{1}
      {4}{{{\color{numb}4}}}{1}
      {5}{{{\color{numb}5}}}{1}
      {6}{{{\color{numb}6}}}{1}
      {7}{{{\color{numb}7}}}{1}
      {8}{{{\color{numb}8}}}{1}
      {9}{{{\color{numb}9}}}{1}
      {\{}{{{\color{delim}{\{}}}}{1}
      {\}}{{{\color{delim}{\}}}}}{1}
      {[}{{{\color{delim}{[}}}}{1}
      {]}{{{\color{delim}{]}}}}{1},
}

\newcommand{\bftab}{\fontseries{b}\selectfont}

\usepackage{xspace}
\usepackage{booktabs}
\usepackage{graphicx}
\def\numquestions{13,722\xspace}

\def\germanquad{GermanQuAD\xspace}
\def\germandpr{GermanDPR\xspace}

\title{GermanQuAD and GermanDPR: \mbox{Improving~Non-English~Question~Answering~and~Passage~Retrieval}}

\author{Timo Möller \and Julian Risch \and Malte Pietsch \\
         deepset GmbH \\ \texttt{\{timo.moeller, julian.risch, malte.pietsch\}@deepset.ai}}

\begin{document}
\maketitle
\begin{abstract}
A major challenge of research on non-English machine reading for question answering (QA) is the lack of annotated datasets.
In this paper, we present \germanquad, a dataset of \numquestions extractive question/answer pairs.
To improve the reproducibility of the dataset creation approach and foster QA research on other languages, we summarize lessons learned and evaluate reformulation of question/answer pairs as a way to speed up the annotation process.
An extractive QA model trained on \germanquad significantly outperforms multilingual models and also shows that machine-translated training data cannot fully substitute hand-annotated training data in the target language.
Finally, we demonstrate the wide range of applications of \germanquad by adapting it to \germandpr, a training dataset for dense passage retrieval (DPR), and train and evaluate the first non-English DPR model.
\end{abstract}

\section{Introduction \& Related Work}
Research on non-English machine reading for question answering (QA) suffers from limited availability of annotated non-English datasets.
With the English SQuAD dataset~\cite{rajpurkar2018know} as a role model, there are only a few resources of similar format, such as the French FQuAD~\cite{dhoffschmidt2020fquad}, the Korean KorQuAD~\cite{lim2019korquad1}, and the Russian SberQuAD~\cite{efimov2020sberquad} datasets.
As an alternative, there are machine-translated datasets for training~\cite{lewis2020mlqa}.
With further improving performance of translation models, machine-translating datasets is a promising direction for the future as it saves the costs of manual annotations.
However, translations can also be a source of errors (some translated answers contain non-negligible errors) and can have a negative impact on the variety of language use that might limit model performance.
%

Another line of research uses multilingual models such as mBERT \cite{devlin2018bert}, XLM \cite{NEURIPS2019_c04c19c2} or XLM-Roberta \cite{conneau2020unsupervised} trained on English QA datasets for zero-shot language transfer to the target domain. 
While these models perform astonishingly well on QA in unseen languages, they do not perform as well as on English QA \cite{lewis2020mlqa,dhoffschmidt2020fquad,lim2019korquad1}.
Furthermore, multilingual models are much larger than their monolingual counterparts, rendering them unsuitable for most production systems where memory consumption and query speed matter.

Besides QA, a highly related task called ``passage retrieval'' has been worked on recently. 
Passage retrievers can be used for document retrieval, but also play an important role when scaling extractive QA models to large document bases (open-domain QA).
Dense Passage Retrieval (DPR)~\cite{karpukhin2020dense} uses two separate, BERT-based encoders~\cite{devlin2018bert} for questions and passages.
In addition to the positive passages (contexts) as contained in existing SQuAD-formatted datasets, the training of DPR leverages hard negative passages that are automatically selected for each query and an optimization trick called in-batch negatives.
In-batch negatives optimization assumes positive passages of one annotated question to be negative passages for a different question.
By adding hard negatives selected from the whole German Wikipedia to \germanquad, we create a training dataset in DPR format, which we call \germandpr. 

Our contributions can be summarized as follows:
First, we present a new dataset of extractive question/answer pairs, \germanquad, and show that training on this dataset achieves a new state of the art for German question answering. 
We also present the first multi-way annotated German test set for a more fair evaluation of QA models.
Second, we evaluate the efficacy of a technique to improve the annotation process, where simple questions are reformulated.
Third, we present a new dataset for dense passage retrieval, \germandpr, and demonstrate another application scenario of \germanquad by converting it to a different format and enriching it with additional data.
This conversion approach could also be used with other QA datasets.
To the best of our knowledge, there is no other non-English DPR model published at the point of writing, presumably due to the lack of suitable training datasets.
Our annotations and the trained models are released under the Creative Commons Attribution 4.0 International License (\href{https://creativecommons.org/licenses/by/4.0/}{CC-BY 4.0}).\footnote{\url{https://deepset.ai/germanquad}}

\section{GermanQuAD Dataset Creation} \label{annotation_dataset}
The creation of \germanquad is inspired by insights from existing datasets as well as our domain labelling experience from several industry projects. 
We combine the strengths of SQuAD, such as high out-of-domain performance, with self-sufficient questions that contain all relevant information for open-domain QA as in the Natural Questions dataset~\cite{naturalquastions_dataset}. 
Our training and test datasets do not overlap like other popular datasets \cite{lewis2020question} and include complex questions that cannot be answered with a single entity or only a few words.

\subsection{Selection of Passages}
\label{sec:selection_of_passeges}
The basis of \germanquad is the German counterpart of English Wikipedia articles' used in SQuAD. 
From the German Wikipedia, we extract each article's text with gensim's WikiExtractor\footnote{\url{https://radimrehurek.com/gensim/corpora/wikicorpus.html}}, split it into passages (paragraphs) and exclude passages that are shorter than 500 characters or mainly contain enumerations. 
We add the title of the German Wikipedia page to each passage.
Since we annotate only one answer per question in the training set and want to reduce the risk of false-negative answers, we limit the length of training passages to 2500 characters. 
For the test set, we keep long passages and only dismiss short ones. 
As some text conversion errors appear especially for embedded latex code, annotators skip passages that are not fully understandable. 

\subsection{Selection of Annotators}
Since we strive to create a high-quality dataset in a controlled environment, we decide against a pure crowd-sourced approach. Instead, we handpick annotators that are familiar with supervised learning, teach them in a 2-hours workshop the use of BERT for QA, give detailed labelling instructions, continuously monitor the annotations, and give feedback wherever necessary. 
We collaborate with students in the area of computational linguistics, computer science, and chemistry and one expert annotator who created more than 20k QA annotations. 
To have a more diverse set of annotated answers in the three-way annotated test dataset, we source around two thousand labels via the crowd service provider Crowd Guru,\footnote{\url{https://www.crowdguru.de/}} which has been selected after careful benchmarking.

\subsection{Annotation Instructions}
Our annotators create questions while reading the passage. 
To prevent a strong lexical overlap of the question and the corresponding passage, a shortcoming of SQuAD \cite{naturalquastions_dataset}, the annotators reformulate the question text with the use of synonyms and altered syntax wherever possible. 
We encourage creation of complex questions that cannot be answered with a single entity. 
Consequently, next to answers that are short (a few words, a single number, or an entity), there are also long answers spanning multiple sentences that cannot be shortened without becoming incomplete or even incorrect. 
A translated example of a short answer to the question: ``What is the EU according to the Federal Constitutional Court?''\footnote{GermanQuAD ID: 57716} is ``association of states''.
An example of a long answer to the question: ``What can be seen on the coat of arms of Hannover?''\footnote{GermanQuAD ID: 56712} is ``a silver wall with two pewter towers on a red ground; in the open gate, under a black portcullis, is a golden shield with a green Mary's flower or shamrock (unexplained); between the towers is a golden lion.''.

The questions are self-sufficient in the sense that no additional information is needed to give an answer. 
To ensure self-sufficient questions, we teach the annotators to classify SQuAD questions into self-sufficient (``When did Israel join the OECD?''\footnote{SQuAD ID: 5725cd6a89a1e219009abef7}) versus incomplete (``What year did Albert die?''\footnote{SQuAD ID: 5723d010f6b826140030fc8e}).
For the complete annotation FAQ, see Appendix~\ref{sec:appendixannotateFAQ}.
Figure~\ref{fig:germanquad} exemplifies the data format.
\begin{figure}
\hspace*{\dimexpr\fboxsep+\fboxrule}%
\begin{minipage}{\dimexpr\linewidth-2\fboxsep-2\fboxrule}
\begin{lstlisting}[language=json,numbers=none]
[{
  "paragraphs": [{"qas": [{
    "question": 
      "Von welchem Gesetzt stammt...?", 
    "id": 51870, 
    "answers": [{
      "answer_id": 53778, 
      "document_id": 43958, 
      "question_id": 51870, 
      "text": "britischen...", 
      "answer_start": 146, 
      "answer_category": "SHORT"
      }]
  }]}],
  "context": "Recht_der...\n\n===...",
  "document_id": 43958
}]
\end{lstlisting}
\end{minipage}
\caption{Sample from the \germanquad dataset.}\label{fig:germanquad}
\end{figure}

\subsection{Annotation Tool}
The annotations are collected with our own QA annotation tool.
It includes basic functionality for extractive QA, such as creating questions while reading a text or answering pre-defined questions on that text. 
Pre-defined questions can be labelled as \emph{no answer possible in the accompanying passage.} 
Answers are marked as a single span within the text and labelled \emph{short,} \emph{long,} \emph{yes,} \emph{no,} and \emph{other.} 
The tool allows importing documents and questions and exporting annotations as xlsx or SQuAD style JSON files. 
It is free-to-use and available online, with an active user community of more than 1k registered users and 80k created annotations.\footnote{\url{https://annotate.deepset.ai/}}
Documentation of using the containerized version for self-hosted labelling is available in the open-source framework Haystack.\footnote{\url{https://github.com/deepset-ai/haystack/tree/master/annotation_tool}}

\subsection{Quality Management}
An expert annotator manually checks the first 300 questions and answers for each annotator and gives detailed feedback.
In weekly group meetings, a random subset of newly created labels is discussed.
We encourage working in groups to debate imprecision in question formulation or answer selection.
We continuously check for lexical overlap between question and passage and correct the annotation style in individual sessions with the corresponding annotator.

\subsection{Addressing Bias}
Of course, we cannot be sure to create a dataset free of bias but by being aware of the potential problems we can strive to overcome at least a few. 
The bias inherent in Wikipedia texts is difficult to overcome when labelling extractive QA on these texts.
We hope the popular wiki pages chosen for our dataset are less biased because they are curated by many people.
The team of annotators includes men and women, and native as well as non-native German speakers. 
We hypothesize questions seeking factual information instead of opinions make the dataset less prone to bias.
Since the gender form is a specific issue in the German language, we encourage to include male and female or gender-neutral forms when formulating questions (``... alle Bürger:innen eingeführt.''\footnote{Training dataset, Question ID: 54576}).

\subsection{Test Set}
We construct a three-way annotated test set including 2204 questions. 
Student annotators formulate all questions and mark the first answer. 
The second answer comes from our expert annotator who also removes questions that are not formulated clearly. 
For the third annotation, we use the crowd annotation provider Crowd Guru. 
Crowd Guru's annotations are checked for positional overlap with existing answers. 
If there is no overlap, we manually check for irregularities and as a result, remove 76 wrong answers. 
The final dataset consists of 2204 questions with $2204\cdot 3-76=6536$ answers.
Inspired by \citet{lewis2020question}, we minimize the overlap of train and test set by comparing normalized\footnote{lowercased, white-space normalized, punctuation and German stop words removed} questions and answers. 
We remove 11 questions from the training set because they also occur in the test set. For 12.3\% of the questions in the test set, one of their answers also occurs in the training set.
This number is low compared to 58\% to 72\% reported by \citet{lewis2020question} for other datasets.

\subsection{Dataset Analysis}
This section describes the created datasets, \germanquad and \germandpr in more detail.
Table~\ref{tab:dataset_germanquad} lists the number of passages, questions, and answers in the training and test set of \germanquad.
Each passage is from a different Wikipedia article and therefore, the number of passages is equivalent to the number of articles.
As the training dataset is one-way annotated, there is exactly one answer per question, resulting in 11518 question/answer pairs.
In contrast to that, the test dataset contains three answers per question:
For 2204 questions, there are 6536 answers.

Statistics of the different question types can give an indication of the diversity and complexity of the questions.
To this end, Table~\ref{tab:question_type} lists question types with examples sorted by frequency in the test set of \germanquad.
Further, it shows the average answer length (and standard deviation) and the model performance with regard to exact match, F1-score, and Top-1-accuracy.

The average answer length is shortest for questions of type \emph{how, how many,} and \emph{when,} which indicates that the answer is less complex, e.g., a year for questions of the type \emph{when}.
Questions of type \emph{what} require the longest answers on average. 
The results of exact match, F1-score, and Top-1-accuracy indicate that questions of type \emph{what} are also among the most difficult, as performance is only lower for questions of type \emph{other}.
Questions of type \emph{when} are answered with highest scores of exact match, F1-score, and third-highest Top-1-accuracy.

\begin{table}
\centering
\begin{tabular}{lrrrr}
\toprule
 & Passages & Questions & Answers\\ 
\midrule
Train & 2540 & 11518 & 11518 \\
Test & 474 & 2204 & 6536 \\
\bottomrule
\end{tabular}
\caption{\germanquad comprises a one-way annotated training set and a three-way annotated test set of German Wikipedia passages (paragraphs). Each passage is from a different article.}\label{tab:dataset_germanquad}
\end{table}

\begin{table*}
\centering
\begin{tabular}{llr S[separate-uncertainty,table-figures-uncertainty=1] rrr}
\toprule
Question & Example & Freq & {Answer Length} & EM & F1 & Top1Acc \\ 
\midrule
Which & Über welche Rechte... & 25.2 &7.3 \pm 9.5 &58.8 & 75.7 & 83.8\\
What & Was ist die EU laut... & 20.3 &11.0 \pm 12.0 &42.4 & 68.3 & 80.1\\
How & Wie kann man auf... & 11.4 &7.0\pm9.3 &49.4 & 70.8 & 80.9\\
When & Wann wurde das erste...  & 11.3 &7.0\pm9.4& 71.8 & 81.0 & 84.7\\
Who & Wer ist Bürgermeister... & 10.5 &8.6\pm10.7& 59.3 & 73.1 & 80.5\\
How many & Wie viele ethnische... &7.0 &7.0\pm9.3& 59.7 & 76.2 & 85.7\\
Where & Wo wird in Estland...  & 5.5 &8.3\pm10.3& 56.2 & 77.3 & 87.6\\
Why & Warum unterscheidet... & 4.9 &7.7\pm9.7& 50.5 & 70.0 & 76.6\\
Other & Woraus werden CDs...  &4.0 &7.4\pm10.0& 30.7 & 64.8 & 79.5\\
\bottomrule
\end{tabular}
\caption{Question types, their relative frequency, and the mean answer length $\pm$ standard deviation in the test set of \germanquad along with our model's performance on each type (in percent).}\label{tab:question_type}
\end{table*}

\section{GermanDPR Dataset Creation}
The SQuAD format is not suitable for training DPR models.
DPR requires additional data and a different data format.
In DPR, each question/answer pair needs to be accompanied not only by the context that contains the answer (positive context) but also by other contexts that are semantically similar to the question but do not contain the answer (hard negative contexts).
We take \germanquad as a starting point and add hard negatives from a dump of the full German Wikipedia.
Following the approach by \citet{karpukhin2020dense}, we use BM25 to find passages in Wikipedia that are most similar to the question but do not contain the answer string.
These passages are then added as hard negatives.
To this end, we split each Wikipedia article into passages as described for the creation of \germanquad in Section~\ref{sec:selection_of_passeges}.
After indexing all passages with Elasticsearch, we use its BM25 implementation to retrieve three hard negatives for each question.
However, we do not use the full \germanquad dataset but only question/answer pairs that have been labelled by the annotators as short answers.
The reason is that hard negatives must not contain the answer, which we take into account by checking for the exact answer string.
This exact string matching cannot catch rephrased versions of the answer, which we found is problematic especially for long answers but rarely for short answers.
Figure~\ref{fig:germandpr} exemplifies the data format.

The final \germandpr dataset comprises 9275 question/answer pairs in the training set and 1025 pairs in the test set.
For each pair, there are one positive context and three hard negative contexts.

\begin{figure}
\hspace*{\dimexpr\fboxsep+\fboxrule}%
\begin{minipage}{\dimexpr\linewidth-2\fboxsep-2\fboxrule}
\begin{lstlisting}[language=json,numbers=none]
[{
  "question": "Wie viele... ?", 
  "answers": ["75 % der..."], 
  "positive_ctxs": [
    {"title": "Gott", 
     "text": "Gott\n\n=== Demografie ===\nEine... "}
  ], 
  "hard_negative_ctxs": [
    {"title": "Christentum", 
     "text": "Christentum\n\n=== Ursprung ... ===\nDie ..."}, 
  ]
}]
\end{lstlisting}
\end{minipage}
\caption{Sample from the \germandpr dataset.}\label{fig:germandpr}
\end{figure}


\section{Experiments \& Results}
The following experiments study the impact of the number of training samples on model performance for extractive QA, the benefits of reformulating previously annotated questions, and the model performance compared to related work and baselines for models trained on \germanquad and \germandpr.

\subsection{Labelling Experiments}
\begin{figure}
  \centering
  \includegraphics[width=1.0\linewidth]{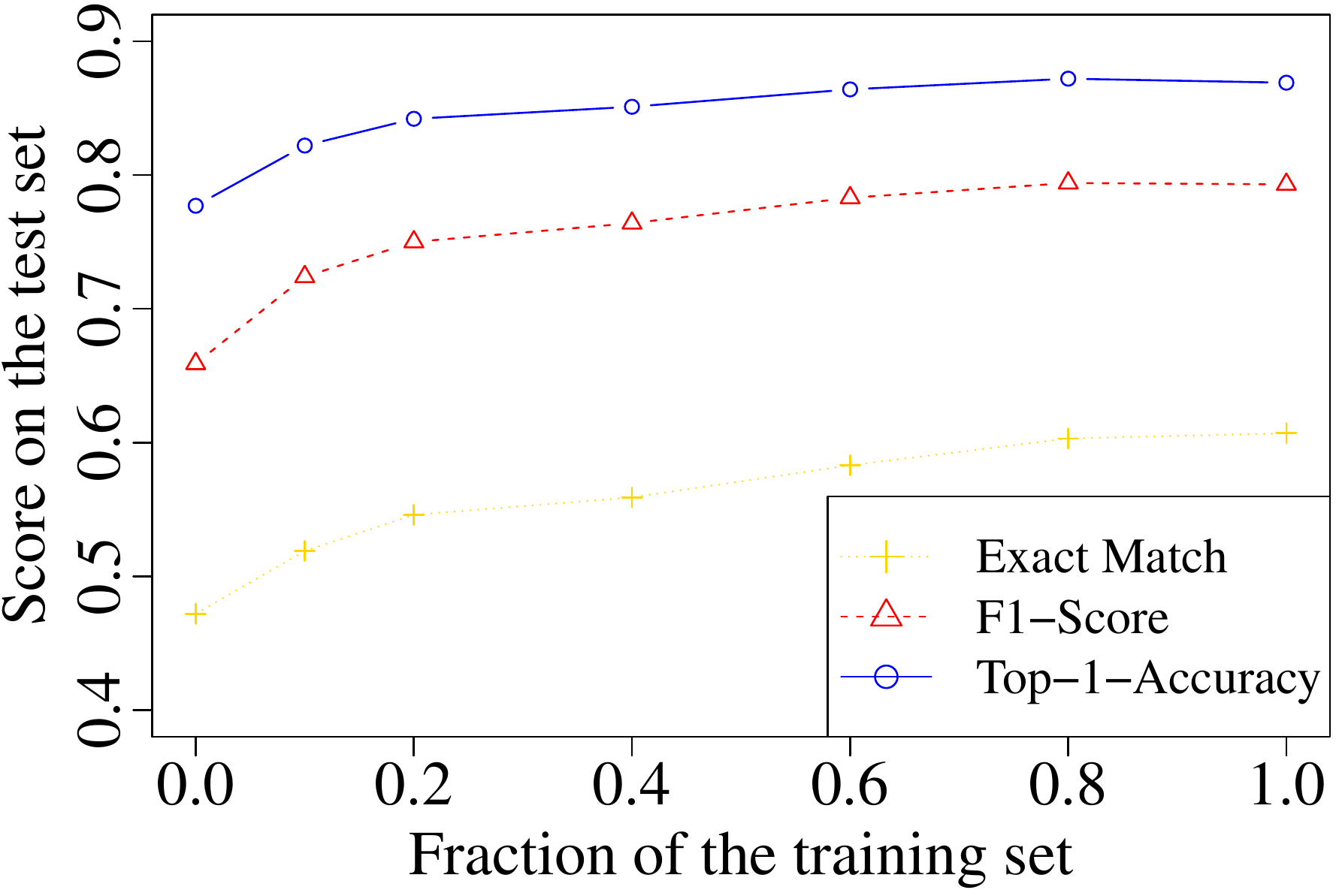}
  \caption{Impact of number of training samples on performance on test set.}\label{fig:num_labels}
\end{figure}

Figure~\ref{fig:num_labels} shows the impact of the number of training samples on the percentage of exact matches, F1-score, and Top-1-accuracy (any overlap of prediction and ground truth) on the \germanquad test set.
For all three metrics, there are no improvements after using more than 80 percent of the training samples.

Inspired by \citet{khashabi2020bang}, we experiment with perturbation of questions.
To this end, annotators created a set of question/answer pairs and we use an existing extractive QA model to predict an answer for each question.
Questions that are correctly answered and where the model has high confidence are simple and presumably not beneficial for model training (high confidence is inferred from large logits of the answer start and end indices).
Instead of completely discarding these simple questions, we reformulate them.
Annotators are much faster at reformulating an existing question compared to formulating a new question and annotating the answer in the context.
The idea of the reformulation is to make the question more difficult while still corresponding to the same answer. 
An example of the original formulation (translated) is: ``Which regions founded the federal state of Canada?'' and the corresponding reformulation is: ``From which regions did the federal state of Canada emerge?''.
The main difference is the changed word order and the use of \emph{emerge} instead of \emph{founded}, which is a less common wording and might result in lower lexical overlap of the question and the context containing the answer.

Table~\ref{tab:reformulation} compares the performance of a model trained on the original question formulations, on a dataset where the 1901 simplest questions are replaced with reformulated questions, and on a joint dataset of both.
It shows that original formulations are not worse than the reformulations for training the model.
However, the reformulations add variety to the training data, which slightly improves model performance when using both the original formulations and the reformulations.

\begin{table}
\centering
\begin{tabular}{lrrr}
\toprule
Data & EM & F1 & Top1Acc \\ 
\midrule
original  & 60.7 & 79.3 & 86.9 \\
reformulations  & 60.2 &\bftab 79.9 & 87.3 \\
orig.+reform.&\bftab 61.5 & 79.8 &\bftab 87.5 \\
\bottomrule
\end{tabular}
\caption{Model performance is almost unchanged if original question formulations are replaced with reformulations in the training data but using \emph{both} slightly boosts performance.}\label{tab:reformulation}
\end{table}

\subsection{QA Performance on \germanquad}
In this section, we compare baseline QA models with models trained on \germanquad. Monolingual baseline models already have German QA capabilities by warm starting them on machine-translated SQuAD supplied along Facebook's MLQA dataset\footnote{\url{https://dl.fbaipublicfiles.com/MLQA/mlqa-translate-train.tar.gz}}. In addition to monolingual models, we use a multilingual XLM-Roberta model~\cite{conneau2020unsupervised} trained on English SQuAD v2.0. 
We continue finetuning the baseline models on the \germanquad training set using the FARM framework.\footnote{\url{https://github.com/deepset-ai/FARM}} 

Next to exact match (EM) and the ratio of overlapping words (F1) metric we also report Top-N-Accuracy scores. 
Top-N-Accuracy is a binary hit or miss score for the first N model predictions.
The value is one if there is any \emph{positional} overlap between the ground-truth answer and the model prediction --- otherwise, it is zero. 
Such a hit or miss score is less prone to details of how answers are annotated (e.g., Answer1: ``19th century'', Answer2: ``middle of 19th century'') and useful in settings without multi-way annotations or where answers can be more than numbers, single entities, or a few words. 
\begin{table*}
\centering
\begin{tabular}{lrrrr}
\toprule
Model & Params & EM & F1 & Top1Acc \\ 
\midrule
Human Baseline & $\approx$100T & 66.4 & 89.5 & 96.4 \\
\midrule
GELECTRA-base-SQuADtranslate & 110M & 47.2 & 65.9 & 77.7 \\
GELECTRA-base-GermanQuAD & 110M & 58.7 & 78.2 & 86.8 \\
XLM-R-base-SQuAD & 270M & 49.1 & 68.6 & 79.9 \\
XLM-R-base-SQuAD-GermanQuAD & 270M & 59.3 & 78.1  & 85.8  \\
\midrule
GELECTRA-large-SQuADtranslate & 335M & 58.5 & 78.8 & 88.7 \\
GELECTRA-large-GermanQuAD & 335M &\bftab 68.6 &\bftab 88.1 &\bftab 93.7 \\
\bottomrule
\end{tabular}
\caption{QA model performance on the three-way annotated \germanquad test set (in percent). Model types and training data are included in the model name. For finetuning XLM-Roberta we use the English SQuAD v2.0 dataset, the GELECTRA models are warm started on the German translation of SQuAD v1.1 and finetuned on \germanquad. The human baseline was computed for the 3-way test set by taking one answer as prediction and the other two as ground truth.}\label{tab:germanquadtest}
\end{table*}

We train our model on \germanquad for two epochs with a learning rate of $3\cdot10^{-5}$ using Adam with default settings, a batch size of 24 and a maximum sequence length of 384 tokens.
Training on \germanquad significantly improves performance on the test set (Table~\ref{tab:germanquadtest}). 
Interestingly, the zero-shot language transfer of \mbox{XLM-R} (XLM-R-base-SQuAD) produces better results than training the monolingual models on translated data (GELECTRA-base-SQuADtranslate). 
When training continues on in-language data though, the situation reverses and the monolingual model (GELECTRA-base-GermanQuAD) becomes slightly better than its multilingual counterpart (XLM-R-base-SQuAD-GermanQuAD). 
In addition to having a better performance, those monolingual models are much smaller compared to their multilingual counterparts (see ``Params'' in Table \ref{tab:germanquadtest}).

\subsection{\germanquad Out-of-Domain Performance}
In order to test how well models trained on \germanquad generalize to other data, 
we evaluate them on the German part of MLQA~\cite{lewis2020mlqa}, XQuAD~\cite{artetxe2020cross}, and a private dataset on financial compliance reports called SFCR.\footnote{SFCR is annotated SQuAD-style and consists of 2386 QA pairs.} 

\begin{table*}
\setlength{\tabcolsep}{3pt}
\centering
\begin{tabular}{lrrrr}
\toprule
Model & MLQA-test & MLQA-dev & XQuAD & SFCR \\ 
\midrule
GELECTRA-base-SQuADtranslate & 37.3/60.1/72.5 & 35.0/59.9/73.1 & 59.2/75.5/82.9 & 23.5/56.3/73.3 \\
GELECTRA-base-GermanQuAD     & 30.9/59.8/76.9 & 27.3/59.4/77.2 & 45.6/70.9/85.0 & 31.9/68.7/82.1 \\
\midrule
GELECTRA-large-SQuADtranslate & \bftab{40.6}\normalfont/\bftab{66.5}\normalfont/78.8 & \bftab{39.1}\normalfont/\bftab{68.9}\normalfont/81.7 & \bftab{62.8}\normalfont/\bftab{81.1}\normalfont/89.2 & 30.4/68.3/84.2 \\
GELECTRA-large-GermanQuAD     & 32.9/64.2/\bftab{81.0} & 31.3/65.9/\bftab{82.8} & 49.1/73.4/\bftab{90.7} & \bftab{39.8}\normalfont/\bftab{78.0}\normalfont/\bftab{91.2} \\
\bottomrule
\end{tabular}
\caption{QA model performance for out-of-domain datasets without the use of training (X-SQuADtranslate) and with (X-\germanquad). The reported metrics are EM/F1/Top-1-Accuracy. The latter is a binary hit or miss based on the position of the best model prediction. }\label{tab:germanquadood}
\end{table*}
Table~\ref{tab:germanquadood} shows the out-of-domain performance on four datasets. 
Training with \germanquad improves Top-1-Accuracy for every dataset, meaning that the model is capable of finding the semantically correct answer more often. 
At the same time, we observe performance drops in EM that are often related to dataset characteristics or metric details. 
Traditional QA metrics, such as Exact Match and F1, are unsuited for datasets that do not have multi-way annotations because they cannot cover all possible correct answers. 
For example, for question ``What did Deng Xiaoping propose in the 1980s?''\footnote{MLQA question:\\ 2ebe96790124698968addfddfb6c6a10a580ca72} we find gold label ``that there should be only one China'' and model prediction ``the reunification of China'' as equally valid. 
F1 and especially EM penalize for different answer styles. 
\germanquad often marks more complete answers (``Lion'' vs ``in star constellation Lion'', ``Maxim Gorki'' versus ``on board the Soviet cruise ship SS Maxim Gorky''), which should not be punished. 
Manual checks for mismatches in Exact Match revealed \germanquad model predictions were often as valid or better as original ground-truth labels.

\subsection{Retrieval Performance on GermanDPR}
On the \germandpr dataset, we train a DPR model that uses two German BERT models~\cite{chan2020germans}, gbert-base, as encoders of the question and the passage.
Pairs of encoded questions and passages are compared with the dot product as similarity function and the number of hard negatives is restricted to 2. 
More hard negatives are included in the dataset and could be used for training but they would also increase the amount of required GPU memory.
Instead, we use in-batch negatives.

For the training, queries are clipped after 32 tokens and passages after 300 tokens.
The learning rate is $10^{-6}$ using Adam, linear
scheduling with warm-up, and dropout rate 0.1.
With a batch size of 40, the model achieves an average in-batch rank of 0.50 after training for 20 epochs and the average rank does not further improve after more epochs.

\begin{figure}[!ht]
  \centering
  \includegraphics[width=1.0\linewidth]{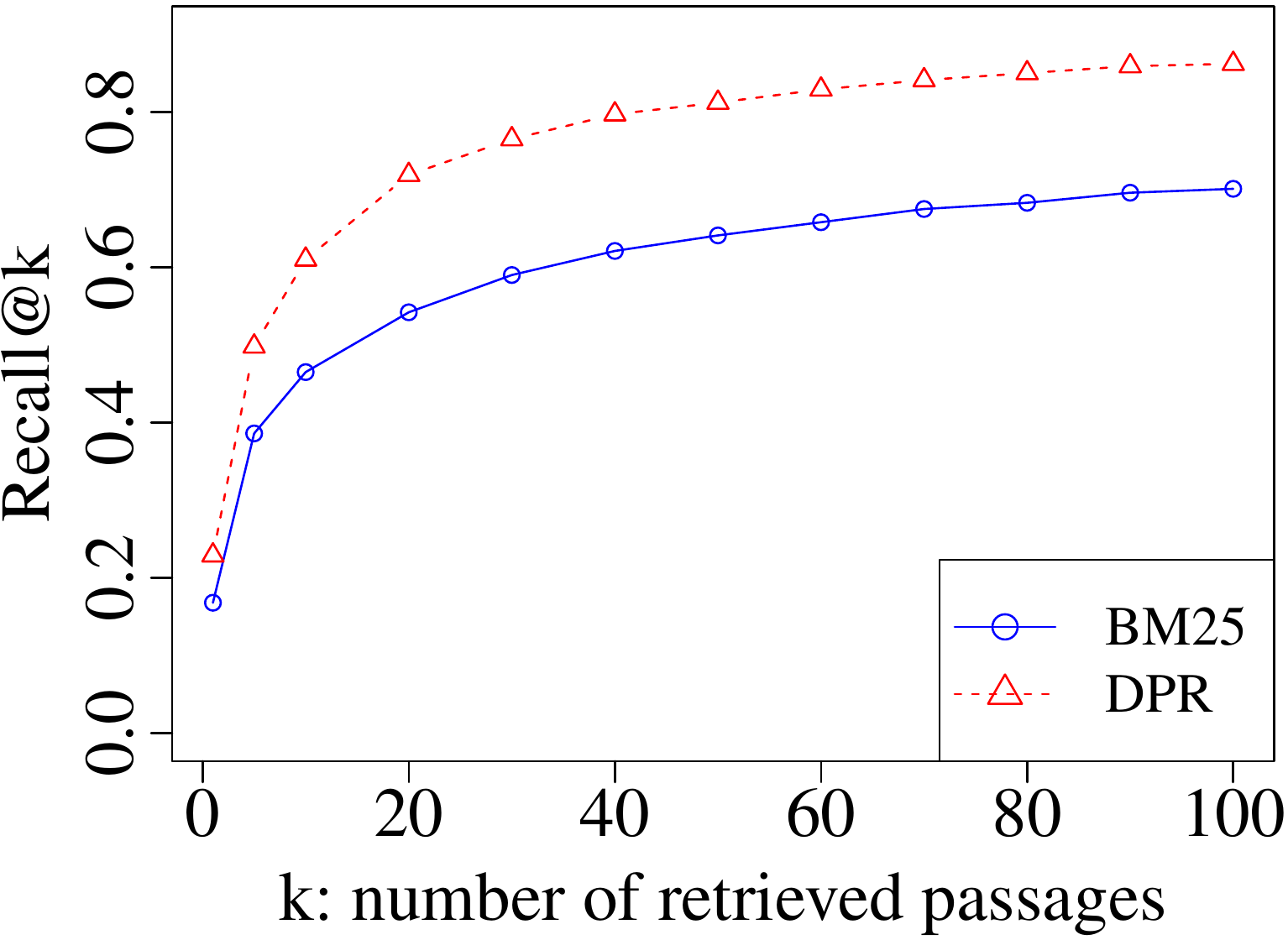}
  \caption{Comparison of BM25 and DPR with regard to recall@k on the full German Wikipedia.}\label{fig:recall_at_k}
\end{figure}

Figure~\ref{fig:recall_at_k} compares the retrieval performance of BM25 and DPR on the full German Wikipedia (2.8 million passages). 
It shows that DPR is outperforming BM25 for different numbers of top-k retrieved results.
While BM25 returned the correct passage within the top 10 results for only 46.54\% of the queries, DPR got the right one for 60.98\% of the queries (Recall@10). 

\section{Conclusion and Future Work}
Only a few resources for non-English question answering and passage retrieval are available.
In this paper, we presented two new datasets, \germanquad and \germandpr, to foster research in this area.
We described the data annotation process in detail so that other researchers can reproduce the approach for other languages.
Our experiments showed that training QA models on \germanquad achieves a new SOTA for German QA and that training DPR models on \germandpr drastically outperforms a BM25 baseline.
A promising path for future work is to research new evaluation metrics for QA.
Especially for longer answers in German, we find that current metrics penalise too much if a predicted answer contains some extra words although being semantically similar to the ground-truth answer.
Future metrics could take into account the semantic similarity of predicted and ground-truth answers.

\section*{Acknowledgements}
We would like to thank Julian Gutsch, Tom Hersperger, Luise Köhler, Iuliia Mozhina, and Justus Peter for supporting the project with data annotation and fruitful discussions on how to improve the labelling process. 
%
\bibliography{emnlp2021}
\bibliographystyle{acl_natbib}
\appendix

\section{Annotation FAQ}
\label{sec:appendixannotateFAQ}
Annotation FAQ for the \germanquad dataset annotated on \href{https://annotate.deepset.ai/}{annotate.deepset.ai}.
\begin{enumerate}
   \item What is a good question?
   \begin{itemize}
     \item A good question is a fact-seeking question that can be answered with an entity (person, organisation, location, etc.) or explanation. A bad question is ambiguous, incomprehensible, dependent on clear false presuppositions, opinion seeking, or not clearly a request for factual information.
     \item The question should ask about information present in the text passage given. It should not be answerable only with additional knowledge or your interpretation. 
     \item Do not copy paste answer text into the question. Good questions do not contain the exact same words as the answer or the context around the answer. The question should be a reformulation with synonyms and in different order as the context of the answer. 
     \item Questions should be very precise natural questions you would ask when you want information from another person. 
   \end{itemize}

   \item How many questions should you ask per text passage?
   \begin{itemize}
     \item Maximally ask 20 questions per passage.
     \item Some text passages are not suited for 20 questions. Do not make up very constructed and complicated questions just to fill up the 20 - move on to the next text.
     \item Try to ask questions covering the whole passage and focus on questions covering important information. Do not only ask questions about a single sentence in that passage.
   \end{itemize}
   
   \item What is a good answer span?
   \begin{itemize}
     \item Always mark whole words. Do not start or end the answer within a word.
     \item For short answers: The answer should be as short and as close to a spoken human answer as possible. Do not include punctuation.
     \item For long answers: Please mark whole sentences with punctuation. The sentences can also pick up parts of the question, or mark even whole text passages. Mark passages only if they are not too large (e.g. not more than 8-10 sentences).
   \end{itemize}
   \balance
   
   \item How do I differentiate long vs short answers?
   \begin{itemize}
     \item If there is a short answer possible you should always select short answer over long answer.
     \item Short precise answers like numbers or a few words are short answers. 
     \item Long answers include lists of possibilities or multiple sentences are needed to answer the question correctly.
   \end{itemize}

   \item How to handle multiple possible answers to a single question?
   \begin{itemize}
     \item As of now there is no functionality to mark multiple answers per single question. 
     \item Workaround: You can add a question with the same text but different answer selection by using the button below the question list (Button reads “custom question”). 
   \end{itemize}

   \item What to do with grammatically wrong or incorrectly spelled questions?
   \begin{itemize}
     \item Include them. When users use the tool and ask questions they will likely contain grammar and spelling errors, too. 
     \item Exception: The question needs to be understandable without reading and interpretation of the corresponding text passage. If you do not understand the question, please mark the question as “I don’t understand the question”.
   \end{itemize}

   \item What to do with text passages that are not properly converted or contain (in part) information that cannot be labelled (e.g. just lists or garbage text)?
   \begin{itemize}
     \item Please do not annotate this text.
     \item You can write down what is missing, or the cause why you cannot label the text + the text number and title.
   \end{itemize}
   
    \item Which browser to use?
   \begin{itemize}
     \item Please use the Chrome browser. The tool is not tested for other browsers.
   \end{itemize}
\end{enumerate}
\end{document}